\documentclass{article}

\PassOptionsToPackage{numbers}{natbib}
\usepackage[preprint]{arxiv_preprint}

\usepackage[utf8]{inputenc} % allow utf-8 input
\usepackage[T1]{fontenc}    % use 8-bit T1 fonts
\usepackage{hyperref}       % hyperlinks
\usepackage{url}            % simple URL typesetting
\usepackage{booktabs}       % professional-quality tables
\usepackage{amsfonts}       % blackboard math symbols
\usepackage{nicefrac}       % compact symbols for 1/2, etc.
\usepackage{microtype}      % microtypography
\usepackage[most]{tcolorbox}
\tcbuselibrary{listings, breakable}

\usepackage[most]{tcolorbox}
\usepackage{xcolor}
\usepackage{amsmath,amssymb}
\usepackage{tabularx}   % For flexible column widths
\usepackage{makecell}   % For line breaks in cells
\usepackage{amsmath, amssymb} % For math symbols
\usepackage{colortbl}   % For colored rows
\usepackage{multirow}
\usepackage{wrapfig}

\usepackage{algorithm}
\usepackage{algorithmic}
\usepackage{amsmath}
\usepackage{amssymb}
\usepackage{graphicx}

\IfFileExists{fontawesome5.sty}{
  \usepackage{fontawesome5}
  \newcommand{\githubicon}{\faGithub}
  \newcommand{\websiteicon}{\faGlobe}
}{
  \newcommand{\githubicon}{\raisebox{0.15ex}{\scriptsize\textbf{GH}}}
  \newcommand{\websiteicon}{\raisebox{0.15ex}{\scriptsize\textbf{WWW}}}
}
\newcommand{\githubcodelink}{%
  {\normalfont\footnotesize
    \githubicon\ \textbf{Code:}~\href{https://github.com/6954717a/Resilience-Eval-eas/}{\nolinkurl{Eas-Resilience-Evaluation}}
    \quad\textbar\quad
    \websiteicon\ \textbf{Website:}~\href{https://6954717a.github.io/resilience_evaluation_web/}{\nolinkurl{Eas-Resilience-Evaluation}}}%
}
\newcommand{\projectresources}{%
  \href{https://anonymous.4open.science/r/EmbodiedAgentSystem\_ResilienceEvaluation-0C3A/}{Code Repository}%
  ~and~\href{https://www.eas-resilience.github.io/}{Project Website}%
}

\definecolor{defpurple}{RGB}{106, 90, 205}   % SlateBlue (学术友好紫)
\definecolor{defpurplebg}{RGB}{245, 242, 255}
\newtcolorbox{defbox}[1]{%
  enhanced,
  breakable,
  colback=green!3,        % 背景浅绿
  colframe=green!50!black,% 边框深绿
  boxrule=0.8pt,
  arc=3mm,                % 圆角
  left=10pt,right=10pt,top=8pt,bottom=8pt,
  sharp corners=southwest,% 可删：让某些角更“论文风”
  fonttitle=\bfseries,
  title={#1},
}
\newtcolorbox{defboxpurple}[1]{
  enhanced,
  breakable,
  colback=defpurplebg,
  colframe=defpurple,
  boxrule=0.8pt,
  arc=3mm,
  left=10pt,right=10pt,top=8pt,bottom=8pt,
  sharp corners=southwest,
  fonttitle=\bfseries,
  title={#1},
}
\newtcblisting{promptbox}[1][]{
  colback=gray!5,
  colframe=gray!50!black,
  listing only,
  breakable, % 允许跨页
  listing options={
    basicstyle=\ttfamily\small,
    breaklines=true,
    columns=fullflexible
  },
  title=#1,
  fonttitle=\bfseries
}

\newtcolorbox{findingbox}[1]{
  colback=gray!3,
  colframe=black!45,
  title=\textbf{#1},
  fonttitle=\small,
  fontupper=\small,
  boxrule=0.45pt,
  arc=1.2mm,
  left=4pt,
  right=4pt,
  top=3pt,
  bottom=3pt,
  before skip=4pt,
  after skip=4pt
}

\definecolor{headergray}{gray}{0.92} % 层级标题背景色
\definecolor{mathcolor}{RGB}{0, 0, 128} % 公式颜色 (深蓝)
\definecolor{signalcolor}{RGB}{80, 80, 80} % 信号文字颜色 (深灰)
\definecolor{groupgray}{gray}{0.95}
\newcolumntype{Y}{>{\raggedright\arraybackslash}X}
\newcolumntype{L}[1]{>{\raggedright\arraybackslash}p{#1}}
\newcolumntype{C}[1]{>{\centering\arraybackslash}p{#1}}

\AtBeginDocument{%
  }

\title{Resilience Matters for Embodied Agents System: New Metrics, Systematic Evaluation, and Optimization}

\author{%
  {\normalfont\normalsize\bfseries
    Yapeng Liu\textsuperscript{\rm 1,\rm 2,\rm 3}\quad
    Yuanzhao Zhai\textsuperscript{\rm 1,\rm 2}\quad
    Xudong Gong\textsuperscript{\rm 1,\rm 2}}\\[-0.10em]
  {\normalfont\normalsize\bfseries
    Dawei Feng\textsuperscript{\rm 1,\rm 2}\quad
    Bo Ding\textsuperscript{\rm 1}\quad
    Lin Wang\textsuperscript{\rm 3}\quad
    Huaimin Wang\textsuperscript{\rm 1,\rm 2}}\\[0.20em]
  {\normalfont\footnotesize
    \textsuperscript{\rm 1}PDL Lab, College of Computer Science and Technology, National University of Defense Technology}\\[-0.10em]
  {\normalfont\footnotesize
    \textsuperscript{\rm 2}State Key Laboratory of Complex \& Critical Software Environment, Changsha 410073, Hunan, China}\\[-0.10em]
  {\normalfont\footnotesize
    \textsuperscript{\rm 3}EmPACT Lab, Nanyang Technological University, Singapore}\\[0.12em]
  {\normalfont\githubcodelink}
}

\begin{document}

\maketitle
\vspace{-0.65em}

\begin{abstract}
% why reliability is critical? 
% 1. Research Question contribution
Embodied Agents System (EAS) are increasingly deployed in open-world physical domains, where reliability directly dictates deployment quality and human-agent trust. However, existing evaluations rely on outcome-centric metrics as success rate or safety scores that collapse diverse execution trajectories into coarse scores, obscuring the dynamic processes underlying agent behavior.
% As Embodied Agents System (EAS) increasingly deployed in physical domains such as autonomous navigation and household assistance, reliable execution becomes critical for physical task completion and collaboration trust. Recent EAS reliability evaluation works typically focus on outcome-centric metrics as success rate or safety-only scores to assess an agent's performance, which collapse diverse execution trajectories into coarse outcomes.
Therefore, they ignore a critical property of EAS -- which we define as the \textbf{Resilience} -- that reflects how EASs \emph{recover}, \emph{stabilize}, and \emph{extend} under perturbations and across iterative updates. The lack of resilience is particularly critical in open-world environments due to continuous unexpected disruptions, thus directly affecting the quality of EAS deployment.
% 2. Metrics contribution
To address this problem, we gain insight from the resilience-engineering concepts to EAS groundings and propose a novel resilience evaluation framework that can be flexibly applied to any EAS. Specifically, we define the first comprehensive resilience metrics suite for EASs system that exposes \textit{Rebound}, \textit{Stability}, and \textit{Graceful Extensibility} across embodied tasks execution, providing a practical grounding for EAS resilience analysis.
% 3. 简化版的resilience evaluation implementation layer
We further implement the resilience evaluation layer that transforms execution process into assessments for diagnosis and optimization. Across 400 household tasks with 10 EAS, we reveal the process-level distinction hidden by outcome metrics, including recovery cost differences among successful episodes ($\Delta C_{\mathrm{rec}}=25.2$), increased instability and task-family degradation.
Metrics-guided optimizations reduce recovery cost and increase stability, graceful extensibility completion, showing the diagnostic effect of resilience evaluation. Our results reveal a trade-off among resilience characteristics, suggesting that a resilient EAS construction should be configured according to deployment-specific requirements.

\end{abstract}

% \keywords{Resilience Engineering, Agents System Test, Embodied Agent System}

\vspace{-0.5cm}
\section{Introduction}
\vspace{-0.2cm}
\label{sec:introduction}

% 【TODO：】
% Introduction这部分的内容我们应该理清：从一个完整的故事出发，来将这三个部分来串起来
% [概念]: 我们希望具身智能系统的韧性，是系统在执行过程中偏离“可接受工作区”后，能否以可控代价回到可接受状态，并在external stress和长期更新中呈现出smooth capability extensibilities。
% 我们提出一个可审计的 resilience measurement layer。它把 EAS 的执行过程分解为恢复、稳定和压力边界三个可计算方面，从而揭示 outcome metrics 无法看到的系统韧性差异。

% [故事实例]: 机器人执行“取物并放置”的长时程任务，原本计划正常推进——中途出现一个轻微但真实的偏差：目标位置Perception偏移、抓取Skill偏移、视觉识别延迟、路径暂时受阻、工具调用失败
% (1) 对 Correction Cost：系统如何通过 retry、replan、rollback、safe fallback 回到可接受范围？回来的代价是多少？
% (2) 对 Stability：若把这个小偏差换成同类的微小变化，系统是否仍能保持相似的恢复模式，而不是出现 plan thrashing、价值振荡、危险抖动？
% (3) 对于 Graceful Extensibility: 在面对于 Stress 的场景下，其退化的特征我们可以如何刻画，是否呈现出脆弱性等特征？

\begin{figure}[t!]
    \centering
    \vspace{-0.5cm}
    \includegraphics[width=\textwidth]{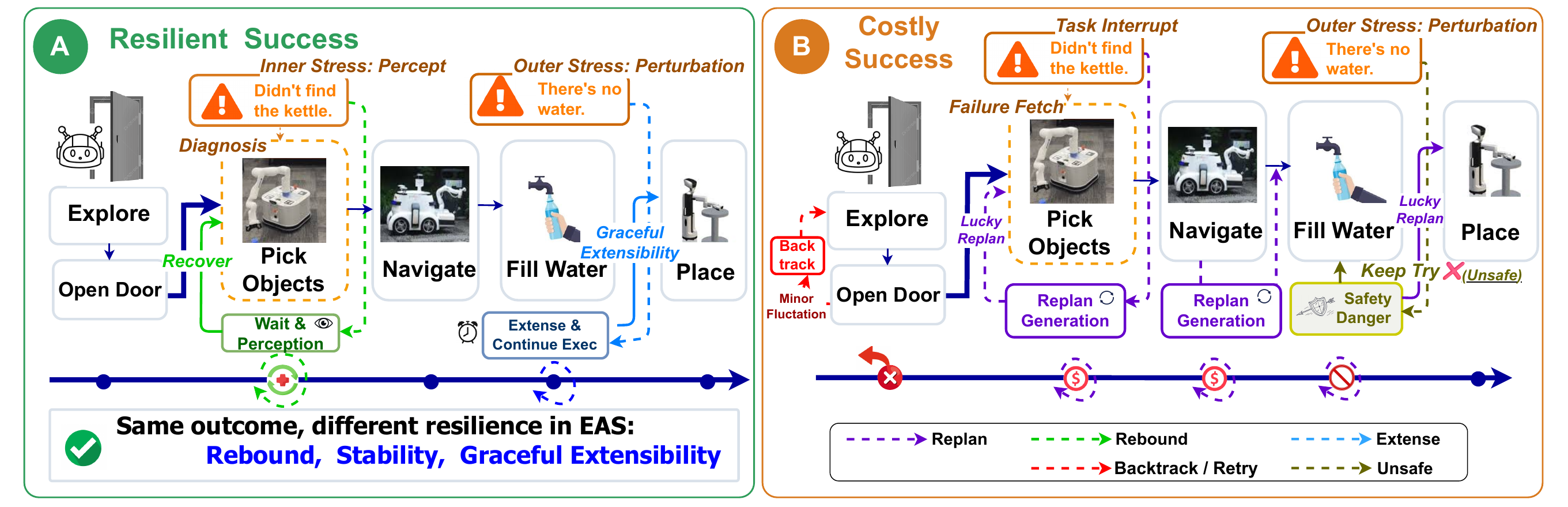}
     \vspace{-0.7cm}
    \caption{Outcome hide process resilience difference in EAS. (A) A resilient success trajectory handles inner and outer stress through resilient mechanisms, (B) A costly success relies on repeated backtrack, replan and unsafe behaviors.}
    \vspace{-0.4cm}
    \label{fig:introduction}
\end{figure}

% 1. 介绍什么是EAS
Embodied Agents System (EAS) shows great potential in domains where physical interaction and long-term task completion are essential, especially for the collaboration domain~\cite{liu2025aligning,buyya2026agentic}. Recent advances in large language models (LLMs) further strengthen EAS capabilities of reasoning and interaction, along with increased complexity in reliability evaluation~\cite{wang2024survey,tan2025towards,chang2024partnr}.
% 2. 这个地方是例子的引入，以场景来引入韧性，说明韧性是重要的（引出 Requirements）
As illustrated in Fig.~\ref{fig:introduction}, two EASs receive the same household instruction and achieve the same outcome, but their thought and execution process can be fundamentally different. This difference matters because EAS operates in dynamic environments where disruptions have continuously emerged~\cite{lu2025bench,fung2025embodied}, indicating that reproducible reliability is critical for physical tasks and collaboration trust during execution~\cite{gao2025current,yokoyama2021success}.
% Since we entrust EASs with physical tasks, they must resiliently handle with disruptions~\cite{gao2025current}. Consequently, \textbf{\emph{Resilience}} evaluation is fundamental for EAS governing utility and reliability, which is revealed through the execution of embodied tasks~\cite{yokoyama2021success}.
% 3. 目前的EAS Evaluation发展进展，同时声明目前依然处于发展不足的状态
However, current EAS evaluation largely remains outcome or safety metrics, such as success rate or aggregated safety triggers~\cite{chang2024partnr,yin2024safeagentbench,li2024embodied,tan2025towards}. They collapse different execution processes into the same score, making brittle success indistinguishable from trustworthy adaptation success~\cite{moskalenko2023resilience,ni2025embodiedarenacomprehensiveunified}. This limitation points to a missing property of EAS -- which we define as the \textbf{\textit{Resilience}} in EAS -- that reflects how EASs \emph{recover}, \emph{stabilize}, and \emph{extend} under stress.

% 4. 我们需要从系统韧性的角度出发，引入到韧性 Evaluation，提出指标 (是说明我们 Research Question 的novel)
To address this problem, we gain insights from the resilience-engineering concepts to EAS groundings. System resilience evaluation captures the intrinsic property of a system, which maintains its acceptable execution and effective recovery when unexpected disturbances happen~\cite{holling1973resilience,avizienis2004basic}. Woods further decomposes system resilience into four capabilities, including rebound, robustness, graceful extensibility under stress, and sustained adaptability~\cite{woods2015four}. 
% Sustained Adaptability is less observable within embodied scenarios where disruption comes often, we therefore focus on the first three aspects~\cite{wied2020conceptualizing,huang2024resilience}.
Specifically, we define the first comprehensive \textbf{Resilience Metrics} for EAS that expose \textit{Rebound}, \textit{Stability}, and \textit{Graceful Extensibility} throughout execution: recovery cost $C_{\mathrm{rec}}$, stability sensitivity $\beta$, stress response mapping $M_f(\lambda)$ and stress capacity$\lambda_f^*$.
% 5. 这里我们要讲 methods, evaluation 的数个层次应该怎么组织
We further implement the \textbf{Resilience Evaluation Layer} that extracts signals from trajectories, monitors, and LLM judge, aggregates them through stage baselines and controlled stress intensities. This layer non-intrusively enables comparison, diagnosis and optimization guidance for EAS resilience evaluation.

% 即使提供了理论支撑，但是如何实现依然是比较难的问题。（这部分是陈述 Implementation 所面对的Dynamic Scenarios 从而导致情况比较难以分析韧性的特征。）
% Although these concepts provide theoretical support, resilience evaluation remains difficult. EAS operate in open-world environments where deviations, recovery behaviors, and acceptable boundaries are often implicit in the execution process. Transforming these qualitative concepts into concrete, measurable evaluation for EAS remains a challenge. 

% 6. 实验结果+Practical Optimization
% 我们围绕 Metrics Validity, Benchmark Results and Practical Optimization 进行了对应的实验。
We conduct experiments on Habitat-Sim with 400 household tasks and 10 representative EAS methods. Focus on EAS resilience Metrics Validity, Benchmark and Practical Optimization, results show that resilience evaluation reveals process-level disparities, benchmarking EAS methods and has practical diagnosis effectiveness.
Among success episodes, physical perturbations increase recovery cost from $11.6$ to $36.3$, semantic perturbations raise instability to $\Delta{\beta}=0.32$, and stress tests expose degradation, recovery cost and perturbations intensities are strongly correlated (Spearman $\rho=0.82$). % These results verifiy the validity of resilience metrics.
For methods, Reflexion and CoPAL achieve comparable SR (56.3\% vs. 55.8\%) but differ by $4.8\times$ in recovery cost, showing that EASs have distinct resilience signatures. There is usually a trade-off in resilience aspects for EAS on our bench. Metrics-guided optimization further improve resilience performance, demonstrating that our evaluation layer is both diagnostic and actionable.

% Contribution 这部分可能还需要加一些数值指标之类的内容。
In summary, our work makes three major contributions.
\begin{enumerate}
  % \item \textbf{EAS Resilience Research Aspects:} We introduce resilience evaluation into EAS and formulate it as a complement to outcome-centric evaluation. We study how EAS recovers, remains stable, and degrades under disturbance, in order to reveal EAS resilience property.
\vspace{-0.1cm}

  \item \textbf{Embodied Resilience Metrics and Evaluation Layer:} We formalize resilience metrics as three aspects: \textit{Rebound}, \textit{Stability}, and \textit{Graceful Extensibility}. Then we complement EAS evaluation with the resilience evaluation layer, a practical grounding for measuring EAS resilience property quantification. 
\vspace{-0.1cm}

  % Reveal Resilience Property
  \item \textbf{Evidence and benchmarking:} We perform a experimental study on 10 representative EAS methods, demonstrating that our metrics successfully distinguish between brittle and resilient behaviors (Sec~\ref{sec:experiments_results_rq1}), we benchmark them, and show their resilience signatures under our resilience evaluation (Sec~\ref{sec:experiments_results_rq2}).
\vspace{-0.1cm}

  \item \textbf{Resilience evaluation provides practical optimization:} We provide quantitative evidence linking specific diagnosis to resilience improvements (Sec~\ref{sec:optimization}), showing that resilience-guided diagnosis can support practical optimization, improving EAS resilience while holding the resilience trade-off design principle.
\end{enumerate}
\vspace{-0.2cm}
% 开源，为了方便复现
% We report hardware, software versions, simulator configurations, and all prompts and tool schemas used, and we release trace logs, judge prompts, and metric analysis scripts in our repository to support full replication of results.

% In this paper, we aim to evaluate EASs encountering disruption, extensibility, recovery, and reorganization, especially when disturbances deviate from what was explicitly anticipated in design-time assumptions. To achieve this, we propose a hierarchical resilience evaluation framework to distinguish different recovery, stability, and extensibility patterns in the practical operation and evolution of EASs, instead of collapsing all performance into an outcome-metric score.
% Specifically, we focus on the situation where the EAS continues to operate within its capacity boundaries and faces the collapse of its own capabilities.

% ========================================================
\vspace{-8pt}
\section{Related Works} % 用来补齐相关方面的知识，避免长篇综述
\vspace{-8pt}
\noindent \textbf{LLM-Enhanced EAS Evaluation: From Outcome to Resilience Metrics.}
% Motivation：(1) 具身智能是与人们日常生活贴合最紧密的领域，所以我们把数据应用验证放在具身智能领域，验证指标可以更贴近生产生活是最需要的部分，也可以借由更真实的数据从而演化迭代;  (2) 具身智能落地；指出被动韧性和主动韧性的区别；具身智能执行效果分析
% (1) 从单纯的规划 -> 到具备自我修正能力的Agent -> 到持续演化的系统 -> 
% (2) 指出评估维度的缺失（Outcome Bias）-> 目前关于 Reliability 的评测目前
% (3) 引入我们自己的系统。
% LLM（逻辑逻辑模型）从静态推理向具身控制的转变，需要动态中断处理机制。SayCan 和 VoxPoser 等早期模型通过值函数和代码策略，将高级指令转化为可行操作。为了解决开环故障，Inner Monologue 和 ReAct 引入了认知反弹循环
LLM-enhanced EAS close the perception-execution loop by taking language-based reasoning into embodied actions generation, and they evolved from instruction grounding to feedback reasoning and self-correction~\cite{ahn2022can,liang2022code,huang2022inner,shinn2023reflexion,skreta2024replan,guo2024doremi}. Recent works~\cite{joublin2024copal,ma2026cyclevla,zhai2025agentevolver,wang2023voyager} further explore adaptation, consistency, and open-ended skill acquisition to build trustworthy EAS. 
However, classical EAS evaluation remains largely outcome-centric as PARTNR emphasizing success metrics~\cite{yokoyama2021success,zhang2025vlabench,cavorsi2023multirobot,or2025mttr}. Under embodied scenarios -- physical irreversibility, partial observability, and dynamic constraints, such outcomes cannot distinguish robust adaptation from brittle success~\cite{liu2025aligning,chang2024partnr,tan2025towards}.

% 这里需要讲其他Reliable Works部分所做的工作内容（一般是添加相应的扰动，然后去区分各个环节的变化）
Focused on EAS reliability evaluation, recent works construct perturbation test to measure safety-aware planning~\cite{yin2024safeagentbench,yang2025embodiedbench,chen2026hazardarena}. They provide reliability diagnostics, but mainly whether an agent violates constraints or approaches a hazardous state in perturbation tests rather than normal process-level execution~\cite{li2024embodied,zhang2026using}. 
We present the resilience metrics to evaluate \emph{EAS Resilience} as an intrinsic property of EAS, measuring how the system behaves from inner stress or perturbations. In this way, we complement EAS evaluation with our resilience evaluation layer to measure how the resilience of EAS exhibits and provide practical optimization guidance.

% \vspace{-0.2cm}
% \vspace{-0.2cm}
% ========================================================
\noindent \textbf{Resilience Concepts Practical Evaluation in EAS Execution.}
% 引入Resilience Engineering Concepts
In systems engineering, resilience emphasizes the capability to endure, adapt to, and recover from disruptions~\cite{prorok2021beyond,cook2021building}. Woods organizes the technical expression of resilience into four concepts: rebound, robustness, graceful extensibility and sustained adaptability~\cite{woods2015four}. This perspective holds that handling open-world complexities relies not only on failure prevention, but on continuous, dynamic adaptation during execution~\cite{hollnagel2018safety}. 
% 传统韧性评估方法，以及他们的局限性
Classical resilience evaluation works evaluate loss severity and recovery latency, but their reliance on exact state matching or crash detection proves insufficient for embodied scenarios~\cite{koopman2016challenges,sharma2018resilience,or2025mttr}. Recently, resilience-relevant evaluating methods for EAS emerged, they evaluate on semantic safety under hazardous instructions and physical cases~\cite{zhu2024earbench,yuan2024r,lu2025bench}. However, these methods mix EAS resilience with adversarial security or perturbation stability. They evaluate the failure under static perturbation, but rarely capture the process-aware dynamics like recovery or extensibility.

% 我们所做的内容
Grounding with resilient EAS construction, we implement the resilience concepts into three resilient aspects. \textit{Rebound} captures the burden of closed-loop recovery; \textit{Stability} captures robustness as decision consistency; and \textit{Graceful Extensibility} captures bounded degradation and remaining operational margin under stress.
Sustained adaptability is treated as a longitudinal extension because it is less directly observable within dynamic EAS execution. Our resilient evaluation realize the practical evaluation layer, providing the validated instrument that reveals the EAS resilience property.
% We validate these phenomena are relevant to the resilience property, locally revealed by resilience metrics. The metrics realize the practical evaluation layer, providing the validated instrument that reveals the EAS resilience property.

% Rebound 考虑的是 "恢复过程相对于局部正常执行的偏离"
% Rebound characterizes how quickly and reliably the system returns from a degraded regime to an acceptable regime. For agentic systems, recovery includes detecting drift, diagnosing causes, and executing corrective actions (re-planning, tool retry, rollback, safe fallback). Rebound metrics measure recovery latency and frequency in trajectories.
% Stability quantifies the persistence of performance under small perturbations and stochasticity. Grounded in sensitivity analysis, it assesses whether the agent maintains concentrated statistics across repeated runs, avoiding brittle mode switches triggered by minor input noise.
% Graceful Extensibility ensures the system retains a conservative utility floor under severe stressors (resource limits or partial observability). We link this to \emph{lower-bound reasoning}, guaranteeing that performance declines predictably rather than collapsing catastrophically~\cite{moskalenko2023resilience}.

% ========================================================
\vspace{-6pt}
\section{The Proposed EAS Resilience Evaluation Framework}
\vspace{-6pt}
\label{sec:metrics}

% Analysis所做的是侧重于整体逻辑框架的定性（做锚点）、理论分析之中的内容; 展示overview的整体逻辑架构，这部分主要说的是我们所做的内容，也就是韧性这一系列Metrics我们所在的指标内容体系
% \begin{figure}[t!]
%     \centering
%     \vspace{-0.2cm}
%     \includegraphics[width=\textwidth]{Images/new_analysis_v1.pdf}
%     \vspace{-10pt}
%     \caption{\textbf{Overview of our EAS Resilience Evaluation framework}. (A) illustrates resilience metrics concept and their computation, and (B) illustrates the resilience evaluation layer pipeline.}
%     % \vspace{-0.6cm}
%     \vspace{-7pt}
%     \label{fig:overview}
% \end{figure}

\noindent \textbf{Overview.} We formalize the Resilience \textbf{Metrics} and Resilience \textbf{Evaluation Layer} for EAS, which goal is to evaluate system resilience from execution dynamics. 
Given an episode artifact $(\tau,\mathcal{L},\mathcal{C},\mathcal{J})$, where $\tau=\{(o_t,a_t,r_t)\}_{t=1}^{T}$ is the execution trajectory, $\mathcal{L}$ is the runtime log, $\mathcal{C}$ is the monitor, and $\mathcal{J}$ is the LLM Judge signal, the evaluation layer maps execution dynamics to three resilience aspects:
\[
\mathcal{M}:(\tau,\mathcal{L},\mathcal{C},\mathcal{J})
\mapsto
(\mathcal{M}_{\text{rebound}},\mathcal{M}_{\text{stability}},\mathcal{M}_{\text{GE}}).
\]

% Our evidence extraction was 
% (1) Collection, Anchor
For episode $i$, let $t$ denote a runtime step. At each step, we record task progress $p_t\in[0,1]$, proposition completion $q_t\in[0,1]$, and execution mode $m_t$. We define a local execution anchor as $a_t := (f, b_t^p, b_t^q, m_t)$, where $f$ is the task family, and $b_t^q$ is the residual proposition load. 
% (2) Stage Baseline
For each anchor, we estimate a local \textbf{Stage Baseline} from clean or reference episodes under the same $a_t$:
\[
\mathcal{N}(a_t):=
\big(
\tau^*(a_t),
W^*_{\mathrm{rem}}(a_t),
\bar{T}^{\mathrm{cog}}(a_t),
\bar{N}^{\mathrm{phy}}(a_t),
\bar{\Delta p}(a_t),
\bar{\Delta q}(a_t),
\bar{r}(a_t)
\big).
\]
Here, $\tau^*(a_t)$ is the nominal cycle time, $W^*_{\mathrm{rem}}(a_t)$ is the estimated remaining work from the anchor, $\bar{T}^{\mathrm{cog}}(a_t)$ and $\bar{N}^{\mathrm{phy}}(a_t)$ summarize cognitive and physical effort, $(\bar{\Delta p}(a_t),\bar{\Delta q}(a_t))$ summarize task progress and proposition completion, $\bar{r}(a_t)$ summarizes nominal risk load. The Stage Baseline provides a common local reference for resilience metrics computation and the reliability of statistical generalization on resilience metrics. 
In parallel, we take LLM-as-a-Judge to densify sparse execution feedback. The LLM judges our process from instruction, state, trajectory, and env feedback, and then produces structured reward for goal progress, rational action, and efficiency; implementation details are available through our \projectresources. 
Based on this shared evidence substrate, we instantiate system resilience from three complementary aspects: \textit{Rebound}, \textit{Stability}, and \textit{Graceful Extensibility} as Table~\ref{tab:resilience_metric_table}. 
\begin{table*}[t]
\centering
\scriptsize
% \footnotesize
\caption{Overview of resilience metrics with aspect, type (\textbf{D}: deterministic outcome, \textbf{S}: statistical, \textbf{A}: analysis, \textbf{T}: time-series), and data source }
\label{tab:resilience_metric_table}
{\footnotesize
\renewcommand{\arraystretch}{1.2}
\setlength{\extrarowheight}{1.8pt}
\setlength{\tabcolsep}{2.6pt}
\begin{tabularx}{\textwidth}{c L{2.7cm} X c c c}
\toprule
\textbf{Sym.} & \textbf{Metric Name} & \textbf{Calculation} & \textbf{Aspect} & \textbf{Type} & \textbf{Data Source} \\
\midrule

$C_{\mathrm{rec}}$ 
& Recovery Cost 
& $C_{\mathrm{rec}}=\frac{1}{W_{\text{rem}}^{*}\left(a_{t_{d}}\right)}\sum_{t_{d}}^{t_{r}} \frac{\Delta \tau_t}{ \tau^{*}\left(a_{t}\right)} \cdot g_{t}^{\text {rec}}$
& Rebound 
& D, S 
& System Logs \\

$\beta$
& Policy Stability
& $\beta = \sup_{z,\rho} \frac{\left| \ell(\pi,\rho(z)) - \ell(\pi,z) \right|}{\|\rho\|}$
& Stability 
& S
& Sensitivity Test \\

$M_f(\lambda)$
& Stress Curve Mapping
& $M_f(\lambda)=\frac{1}{K}\sum_{j=1}^{K}\tilde{r}_{(j),f}(\lambda)-1$
& Extensibility
& D, S
& Stress Test \\

$\lambda_f^*$
& Stress Capacity
& $\lambda_f^*=\sup\{\lambda\in[0,1]\mid M_f(\lambda)\ge 0\}$
& Extensibility
& S, A
& Stress Grid \\

\bottomrule
\end{tabularx}
}
\vspace{-0.7cm}
\end{table*}

\vspace{-0.3cm}
\subsection{Metrics Design}
\vspace{-0.3cm}
\label{sec:metrics_design}
% 调整公式间距
\setlength{\abovedisplayskip}{3pt} % 默认通常是 10pt-12pt，改小
\setlength{\belowdisplayskip}{3pt}

% (1) Rebound:说明Rebound的指标主要是 Rebound Cost: Rebound具体指的是什么过程，为什么要采用 Rebound Cost作为指标                                           
\noindent \textbf{1) Rebound Metrics: Recovery Cost to Availability.}
% (1) 声明 C_rec 的逻辑，为什么我们要分成 认知恢复、物理恢复以及状态负债 这三个部分？他们所代表的含义是什么
% 【Attachments】：这里关于Rebound的详细过程，引到Appendix中的描述。
Rebound measures how much extra work the agent must expend to return the acceptable execution after disruption. We use the \textbf{Recovery Cost} as the primary indicator, revealing the efficient rebound and brute-force rebound.
Detailed implementation is available through our \projectresources.
% (2) 这里我们需要声明 tr, td 的定义（这个地方还要说明Windows等设计吗？或者我们直接说采用滑动窗口的形式完成了匹配）
For a rebound window $[t_d,t_r]$, $t_d$ is the first step at which execution departs from the nominal continuation under the local anchor $a_t$, and $t_r$ is the first step at which the task re-enters the acceptable execution. 
Execution \emph{StageBaseline} $\mathcal N(a_t)$ (execution local reference, refer to Sec~\ref{sec:metrics_pipeline}) provides the nominal local reference, including the remaining nominal work $W^*_{\mathrm{rem}}(a_t)$, and the expected local execution statistics.

% (3) 详细地说明这三部分的具体实现是什么样的，计算过程主要用少量的文字和对应尽量简洁、目的明确的公式来说明，包括各个部分的计算过程、统计意义上的偏差、根据偏差的马哈拉诺比斯距离计算、怎么统一到 C_rec的。
% Rebound process is not a single delay term. It is the joint effect of \emph{extra reasoning}, \emph{extra actuation}, and \emph{extra task debt}.
We decompose recovery cost into three aligned parts: \textbf{Cognitive recovery}, \textbf{Physical recovery}, and \textbf{State debt} in Eq.~\ref{eq:rebound_triplet_simple}.
We aggregate their excess variables as
\begin{equation}
\label{eq:rebound_triplet_simple}
\mathbf{z}_t^{\mathrm{cog}}
=
\frac{(T_t^{\mathrm{cog}}-\bar T^{\mathrm{cog}}(a_t))_+}{\bar T^{\mathrm{cog}}(a_t)+\epsilon},
\quad
\mathbf{z}_t^{\mathrm{phy}}
=
\frac{(N_t^{\mathrm{phy}}-\bar N^{\mathrm{phy}}(a_t))_+}{\bar N^{\mathrm{phy}}(a_t)+\epsilon},
\quad
\mathbf{z}_t^{\mathrm{deb}}
=
\frac{(r_t-\bar r(a_t))_+}{\bar r(a_t)+\epsilon}.
\end{equation}
where $\mathbf{z}_t^{\mathrm{cog}}$ measures excess planning, perception effort; $\mathbf{z}_t^{\mathrm{phy}}$ measures excess navigation, retry effort; and $\mathbf{z}_t^{\mathrm{deb}}$ measures lagged progress leakage, risk load. $(\cdot)_+$ indicates the ReLU function, we record the excess costs while discarding pre-complete~\cite{agarap2018deep}. Implementation details and illustrative calculations are available through our \projectresources.
% (4) task-Family Level说明怎么在统计意义上进行聚合
To make the three channels comparable under the same anchor-conditioned regime, we convert each local excess vector into a covariance-adjusted deviation as Mahalanobis Distance\cite{mahalanobis2018generalized}:
\begin{equation}
\label{eq:mahalanobis_rebound}
g_t^{u}
=
\sqrt{
\mathbf{z}_t^{u\top}
\bigl(\Sigma_u(a_t)+\eta I\bigr)^{-1}
\mathbf{z}_t^{u}
},
\qquad
u\in\{\mathrm{cog},\mathrm{phy},\mathrm{deb}\},
\end{equation}
where $\Sigma_u(a_t)$ is the covariance estimated from standard stage baselines, $\eta I$ is the regularizer. This step removes scale inconsistency and accounts for correlation within each recovery channel. We then aggregate the three channel deviations into recovery intensity $g_t^{\mathrm{rec}}=\sqrt{\frac{1}{U}\sum_{u}^U(g_t^{u})^2}$.
Finally, for episode $i$, the \textit{Rebound Cost} is defined as
\begin{equation}
\label{eq:crec_episode}
C_{\mathrm{rec}}^{(i)}
=
\frac{1}{W^*_{\mathrm{rem}}(a_{t_d})+\epsilon}
\sum_{t=t_d}^{t_r}
\frac{\Delta\tau_t}{\tau^*(a_t)+\epsilon}
\cdot g_t^{\mathrm{rec}}.
\end{equation}
This quantity measures the total extra work required to restore acceptable execution after disruption. Normalization by $W^*_{\mathrm{rem}}(a_{t_d})$ ensures that recovery is judged against the remaining nominal work at the disturbance point. We aggregate rebound signatures using the statistical procedures documented in our \projectresources.

\noindent \textbf{2) Stability Metrics: Execution Fluctuations.}
Stability quantifies the system's ability to remain \emph{consistent} under stresses, whether arising from stochasticity (Execution), perturbations (Sensitivity), or self-evolution. It is reflected in execution as frequent replanning, unstable progress, and abrupt shifts in the progress. We take \emph{Policy Stability}($\beta$) as the primary indicator, because it directly measures whether small changes in task instruction or execution condition amplify into process fluctuations. We describe how $\beta$ is computed from execution representations; implementation details, including the value-network design, are available through our \projectresources.

We use the value function $V(\cdot)$ for each EAS execution representation, which is used to diagnose execution progress changes trend. Given the encoded state $s_t$, the value function is
\begin{equation}
\label{eq:value_function_stab}
V(s_t)
=
\mathbb{E}_{\pi}\!\left[
\sum_{u=t}^{T}\gamma^{u-t}\tilde r_u
\mid s_t
\right],
\end{equation}
where $\pi$ is the agent policy and $\tilde r_u$ is the shaped reward by LLM judge and $W^*_{\mathrm{rem}}$. For each state transition, we compute the one-step temporal-difference residual $\delta_t = \tilde r_t + \gamma (1-d_t) V(s_{t+1}) - V(s_t)$, where $d_t$ is the termination signal.
We further compute the generalized advantage estimate $A_t$ as an exponentially decayed accumulation of TD residuals: $A_t=\sum_{\ell=0}^{T-t}(\gamma\lambda)^\ell\delta_{t+\ell}$.
% These representations are sufficient to measure the execution oscillation over the whole trajectory. 
Then our resilience loss for shape reward update is
\begin{equation}
\label{eq:osc_loss}
\ell(\pi,z_i)
=
\frac{1}{T_i}
\sum_{t=1}^{T_i}
\sqrt{
\mathbf{x}_t^{\top}
\left(
\Sigma(a_t)+\eta I
\right)^{-1}
\mathbf{x}_t
},
\end{equation}
where $z_i$ denotes an episode, $\Sigma(a_t)$ is the standard stage baselines covariance, and $\eta I$ is regularizer. This loss increases when the trajectory $\mathbf{x}_t$ contains large value variance, TD spikes, and advantages. We define \emph{$\beta$-Stability} as this perturbations sensitivity to execution progress:
\begin{equation}
\label{eq:beta_stability}
\beta_f
=
\sup_{z\in\mathcal{Z}_f,\rho\in\mathcal{P}_f}
\frac{
\left|
\ell(\pi,\rho(z))
-
\ell(\pi,z)
\right|
}{
\|\rho\|+\epsilon
}.
\end{equation}
Here, $\mathcal{Z}_f$ is the episode set from task family $f$, $\mathcal{P}_f$ is the controlled perturbations set. A low $\beta_f$ means that similar conditions lead to similar results, while high $\beta_f$ means more replanning, value fluctuation, and progress instability.

% \vspace{-0.3cm}
% \vspace{-0.3cm}
\noindent \textbf{3) Graceful Extensibility Metrics: Stress Response.}
% 【Appendix】: 怎么设置的压力强度，λ和Intensities是如何设计的
Graceful Extensibility measures how an EAS performance extends as stress increases. It ensures that when stressors exceed the system's adaptive capacity, performance declines \emph{gracefully} and predictably rather than collapsing catastrophically. This is a dynamic stress-response, revealing the resilience property of EAS. We therefore model Graceful Extensibility as a stress-response mapping $\lambda \mapsto M_f(\lambda)$, where $\lambda$ is the stress severity and $M_f(\lambda)$ is the operational margin.

Let $\mathcal{T}_{\lambda}$ denote a stress operator with severity $\lambda\in[0,1]$, which is related to perturbations and EAS self-evolvement. In our resilience evaluation, we unify external perturbation and internal evolution through the same stress-response formulation, as documented in our \projectresources. 
For an episode $i$ from task family $f$ under stress $\mathcal{T}_{\lambda}$, we first define a \textbf{\emph{hard}} boundary monitor:
\begin{equation}
\label{eq:episode_margin_no_version}
m_i(\lambda)
=
\min
\left\{
\frac{p_{\mathcal{T},i}(\lambda)}{\tau_p},
\frac{q_{\mathcal{T},i}(\lambda)}{\tau_q},
\frac{\tau_{\mathrm{rec}}}{C_{\mathrm{rec},i}(\lambda)+\epsilon},
\frac{\tau_{\mathrm{stab}}}{\beta_{\mathrm{stab},i}(\lambda)+\epsilon},
\frac{\tau_{\mathrm{safe}}}{U_i(\lambda)+\epsilon}
\right\}
-1 .
\end{equation}
where $U_i$ is the constraints violation load. The thresholds define the minimal acceptable task contract, $m_i(\lambda)<0$ means that the contract is violated.

After accumulation of stage baselines from the task family $f$, we aggregate the statistical extensibility of the operation. For compact notation, let $Q_{\alpha}^{f}[X](\lambda) \triangleq Q_{\alpha}\!\left(X(\lambda)\mid f\right)$, where $Q_{\alpha}^{f}[X](\lambda)$ denotes the $\alpha$-quantile of variable $X$ over episodes from family $f$ under stress $\lambda$. We construct the statistical extensibility vector 
\begin{equation}
\label{eq:ge_ratio_vector}
\resizebox{0.95\columnwidth}{!}{%
$
\displaystyle
\mathbf r_f(\lambda)
=
\left[
\frac{Q_{\alpha}^{f}[p_T](\lambda)}{\tau_p}, \,
\frac{Q_{\alpha}^{f}[q_T](\lambda)}{\tau_q}, \,
\frac{\tau_{\mathrm{rec}}}{Q_{1-\alpha}^{f}[C_{\mathrm{rec}}](\lambda)+\epsilon}, \,
\frac{\tau_{\mathrm{stab}}}{Q_{1-\alpha}^{f}[S_{\mathrm{stab}}](\lambda)+\epsilon}, \,
\frac{\tau_{\mathrm{safe}}}{Q_{1-\alpha}^{f}[U](\lambda)+\epsilon}
\right]^{\top}.
$
}
\end{equation}
We take lower and upper quantiles for benefit variables and cost variables, which makes our metrics sensitive to graceful extensibility.
Let $r_{(1),f}(\lambda)\le \cdots \le r_{(K_{\textrm{max}}),f}(\lambda)$ be the sorted entries of $\mathbf r_f(\lambda)$. The statistical operational margin is defined as
\begin{equation}
\label{eq:statistical_operational_margin}
M_f(\lambda) = \frac{1}{K} \sum_{j=1}^{K} r_{(j),f}(\lambda)-1, \qquad 1<K\le K_{\textrm{max}} .
\end{equation}
This bottom-$K$ aggregation preserves the semantics of extensibility, but avoids the rigid behavior of a minimum optimization. 
% 这部分很重要，说明我们是一个映射
For practical diagnosis, dynamic \emph{\textbf{mapping}} between $M_f(\lambda)$ extensibility and execution performance is an important indicator to analyze EAS resilience. Further, the quantitative Graceful Extensibility indicator is the largest stress level where the EAS remains acceptable execution:
\begin{equation}
\label{eq:ge_capacity_simple}
\lambda_f^*
=
\sup
\left\{
\lambda\in[0,1]
\mid
M_f(\lambda)\ge 0
\right\}.
\end{equation}
We call $\lambda_f^*$ the \emph{stress capacity}. It gives the operational boundary of task family $f$: the maximum stress severity at which the system can still preserve its positive task progress in task family $f$.

\vspace{-0.3cm}
   \subsection{Resilience Evaluation Benchmark}
\vspace{-0.3cm}
\label{sec:metrics_pipeline}

% 这部分 Construction 中，我们把 Resilience Metrics 变成落实到在 Embodied Task Execution 中可以实际运行的 Non-Intrusive Measurement Instruments
\vspace{-0.3cm}
\noindent \paragraph{The Resilience Evaluation Layer Construction}
We construct the Resilience Evaluation Layer as a non-intrusive evaluation instrument that complements the EAS reliability evaluation process. We present our resilience evaluation pipeline in Alg~\ref{alg:metrics_pipeline}, it attaches probes during the EAS execution, then computes collected signals to resilience metrics.
This design allows resilience to be evaluated as an execution property rather than an additional task objective.

\begin{algorithm}[t]
\caption{Resilience Metrics Evaluation Pipeline}
\label{alg:metrics_pipeline}
\small
\begin{algorithmic}[1]
\REQUIRE Task family set $\mathcal{F}$; episode artifacts 
$\{(\tau_i,\mathcal{L}_i,\mathcal{C}_i,\mathcal{J}_i)\}_{i=1}^{N}$; StageBaseline estimation; 
perturbation set $\mathcal{P}_f$; stress grid $\Lambda=\{\lambda_j\}_{j=1}^{m}$; 
fixed thresholds and windows.
% evolve versions $\pi^{(0:K)}$.
\ENSURE $\mathcal{M}_{\mathrm{rebound}}$, $\mathcal{M}_{\mathrm{stability}}$, $\mathcal{M}_{\mathrm{GE}}$
\STATE Estimate Stage Baselines $\mathcal{N}(a)$ from executions.
\FOR{each task family $f\in\mathcal{F}$}
    % Stress Level
    \FOR{each stress level $\lambda\in\Lambda$}
        % Episode
        \FOR{each episode $i$, artifact $(\tau_i,\mathcal{L}_i,\mathcal{C}_i,\mathcal{J}_i)$}
          \STATE Extract stepwise signals: progress $p_t, q_t$, execution mode $m_t$, execution $\mathcal{L}_i$, critics $\mathcal{C}_i, \mathcal{J}_i$.
          \STATE Assign local execution anchors $a_t=(f,b_t^p,b_t^q,m_t)$.
          \IF{Detect \textbf{Rebound} window $[t_d,t_r]$}
            \STATE Compute $\mathbf{z}_t^{\mathrm{cog}}$, $\mathbf{z}_t^{\mathrm{phy}}$ and $\mathbf{z}_t^{\mathrm{deb}}$.
            \STATE Compute rebound cost $C_{\mathrm{rec}}^{(i)}$ from $g_t^{\mathrm{rec}}$.
            \STATE Return to acceptable $\mathcal{N}(a_t)$.
          \ENDIF 
          \STATE Collect \textbf{Stability} values $V(s_t)$, TD residuals $\delta_t$, and GAE advantages $A_t$.
          \STATE Aggregate to vector $\mathbf{x}_t$ and $\beta_i$.
          \STATE Backpropagation Critic $\ell_{\mathrm{stab}}(\pi,z_i)$ and Stage Baselines $\mathcal{N}_f(a)$.
        \ENDFOR
        
        \STATE Retrieve stressed executions under $\mathcal{T}_{\lambda}$.
        \STATE Compute hard boundary monitors $m_i(\lambda)$ in family $f$.
        \STATE Aggregate episode statistics $\mathbf{r}_f(\lambda)$.
        \STATE Compute stress mapping: $M_f(\lambda) \leftarrow \frac{1}{K}\sum_{j=1}^{K} r_{(j),f}(\lambda)-1$.
    \ENDFOR
    \STATE Compute Graceful Extensibility stress capacity: $\lambda_f^* \leftarrow \sup\{\lambda\in\Lambda \mid M_f(\lambda)\ge 0\}$.
    \STATE Update Stage Baseline $\mathcal{N}$ and Statistical Aggregation Metrics
\ENDFOR
\RETURN $C_{\mathrm{rec}}$, $\beta$, $M_f$, $\lambda_f^*$, trajectory.
\end{algorithmic}
% \vspace{-10pt}
\end{algorithm}
% \noindent \textbf{Resilience Metrics Computation.}
% The structured trace is projected into the three metric channels defined in Sec.~\ref{sec:metrics_design}. 
% Recovery Cost is computed from disturbance-recovery windows, $\beta$-Stability from perturbation-sensitive execution fluctuations, and Stress-response Mapping from stressed executions over $\lambda$. When normalization is required, the computation refers to local stage baselines to compare effort, progress, and risk under the same execution context. In this way, the layer turns outcome-centric evaluation into a trace-grounded resilience profile.

\vspace{-0.5cm}
\subsubsection{Resilience EAS Testbed Setup} 
\vspace{-0.2cm}
We conduct experiments on \textit{Habitat-Sim 3.0}~\cite{puig2023habitat} with PARTNR-style~\cite{chang2024partnr} collaborative household tasks, including navigation, object manipulation, and object transportation scenarios. To expose different EAS resilience, we introduce controlled perturbations over physical states, semantic instructions, object states, and stress severity. The full perturbation specification, evaluated EAS methods, and dataset setting are documented in our \projectresources.

% In parallel, we construct stress tests including perturbations and execution-level disruptions, they change the difficulty of grounding or execution while preserving the main instruction.
% All experiments are implemented on our Resilience Evaluation Layer testbed, which couples embodied execution with rebound, and LLM-as-a-Judge mechanism. Experiments conducted on NVIDIA A800 GPUs, take the VLLM reasoning method~\cite{kwon2023efficient}, each method was evaluated for 3 epochs, with each epoch approximately consuming 44 GPU hours. 

% We evaluated the resilience evaluation layer as a process-level measurement instrument for EAS, focus on two questions: \textbf{Exp-Q1 Instrument-Metrics Validity: } Do the resilience metrics exhibit construct validity and reliability?
% \textbf{Exp-Q2 EAS Resilience Benchmarking Efficacy: } How do state-of-the-art EAS methods perform under resilience evaluation?
% % \textbf{Exp-Q3 Practical Effectiveness: } Can resilience diagnosis provide practical and effective signals for EAS improvement? Does resilience generalize across different LLM backbones? 

% RQ1: 指标有效性验证
\vspace{-0.3cm}
% \subsubsection{Exp-Q1: Resilience Metrics as Validated Measurement Instruments}
\subsubsection{Resilience Metrics Validation: Execution Differences Hidden by Outcomes }
\vspace{-0.3cm}
\label{sec:experiments_results_rq1}

% Exp-Q1 evaluates the resilience metrics $(\mathcal{M}_{\mathrm{rebound}},\mathcal{M}_{\mathrm{stability}},\mathcal{M}_{\mathrm{GE}})$ capture distinct resilience property that is hidden by outcome-centric evaluation. We validate the Rebound, Stability, and Graceful Extensibility aspects, while keeping them under the measurement-instrument question.
We first validate whether resilience metrics $(\mathcal{M}_{\mathrm{rebound}},\mathcal{M}_{\mathrm{stability}},\mathcal{M}_{\mathrm{GE}})$ capture distinct resilience property hidden by outcome-centric evaluation, and verify them as the measurement instruments.

\begin{wrapfigure}{l}{0.45\textwidth} % {r}表示靠右，0.45\textwidth是图片区域的宽度
    \centering
    \includegraphics[width=\linewidth]{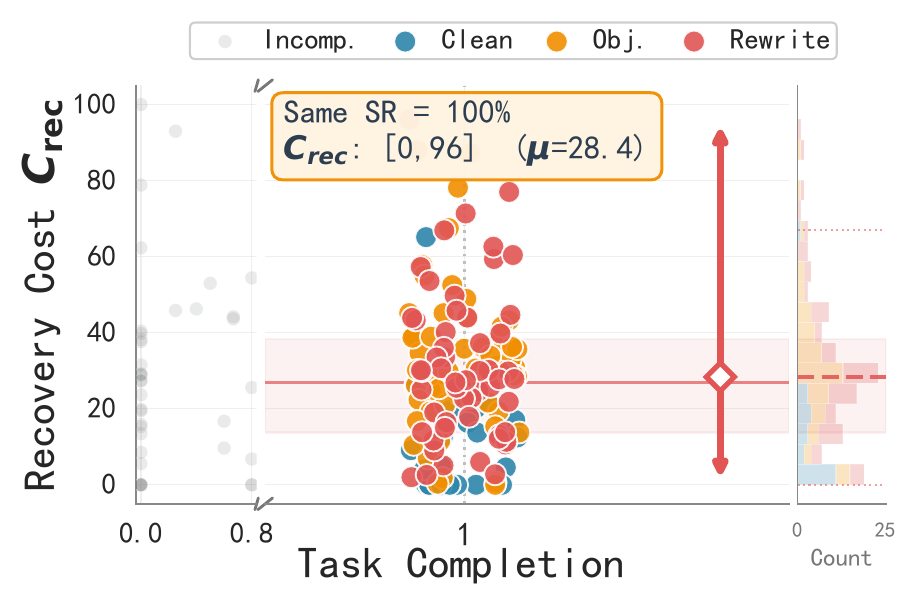} 
    \vspace{-0.7cm}
    \caption{Different $C_{\mathrm{rec}}$ under outcomes.}
    \vspace{-0.9cm}
    \label{fig:metrics_construct_validity}
\end{wrapfigure}

\vspace{-0.7cm}
% Resilience Metrics 具有 Discriminative Difference 的能力
% \noindent \paragraph{Resilience metrics disperse differences in same outcome.} Resilience metrics reveal significant performance differences that both SR and safety scores fail to capture, demonstrating discrimination capability for EAS with similar success rates and safety scores. We take $C_{\mathrm{rec}}$ as an example, Fig~\ref{fig:metrics_construct_validity} shows the dispersion among success episodes with "the same outcome, different resilience".
\noindent \paragraph{Resilience metrics disperse differences in same outcome.} Resilience metrics reveal significant performance differences that both SR and safety scores fail to capture, demonstrating discrimination capability for EAS with similar success rates and safety scores. We take $C_{\mathrm{rec}}$ as an example, Fig.~\ref{fig:metrics_construct_validity} illustrates "the same outcome, different resilience" effect.

\begin{figure}[h]
    \centering
    \includegraphics[width=\textwidth]{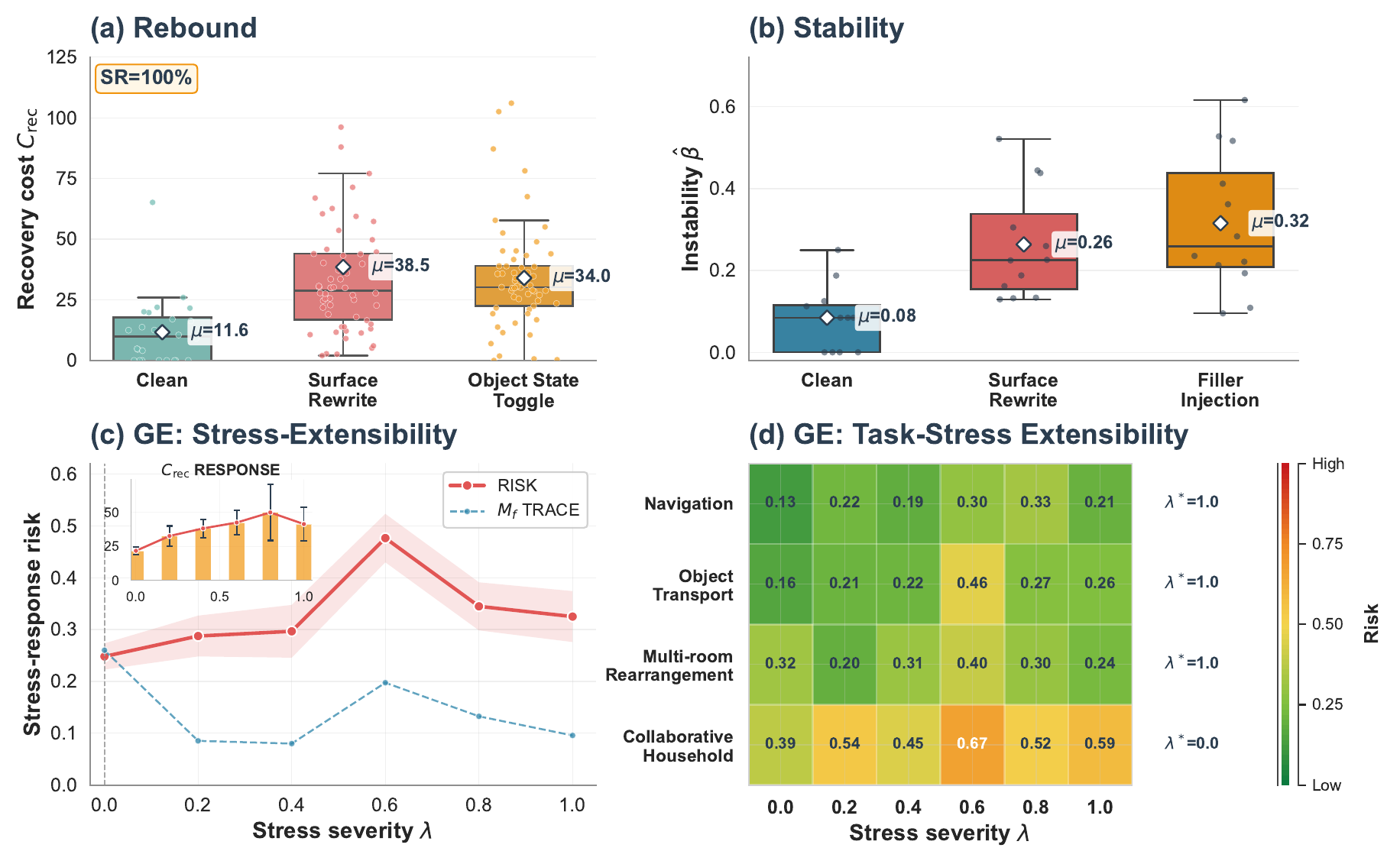}
    % \vspace{-0.6cm}
    \vspace{-0.8cm}
    \caption{Resilience metrics construct validity results (\textit{zoom in for better view}).}
    \vspace{-0.4cm}
    \label{fig:expq1_metric_validity_map}
\end{figure}

\vspace{-0.4cm}
% 韧性构造效度，展示韧性指标与传统指标的相关度、区分度；重点在于区分度
\noindent \paragraph{Resilience Metrics Construct Validity and Sensitivity.}
Fig.~\ref{fig:expq1_metric_validity_map} further confirms the construct validity of the three metric families. Rebound $C_{\mathrm{rec}}$ separates executions by recovery burden: perturbations increase recovery cost from $11.6$ in clean episodes to $36.3$ in average, with a significant distributional shift (Mann--Whitney $p<0.001$). Stability $\beta$ captures sensitivity to semantic-preserving instruction changes: semantic perturbations inflate instability to $\beta=0.32$, mainly driven by divergent action generation instability ($\beta_{\mathrm{out}}=0.42$). Graceful Extensibility characterizes degradation as a stress-response curve rather than a binary failure event: as stress severity increases over $\lambda\in[0.2,1.0]$, the relative margin $\Delta M$ decreases while recovery cost rises, with strong correlation (Spearman $\rho=0.82$). These results show that the metrics measure complementary resilience properties and are not reducible to SR, safety, or completion alone. 

In addition, our \projectresources{} provide analyses of the statistical reliability and sensitivity of the resilience metrics. These analyses indicate that resilience metrics are related to EAS inherent property, which maintains the consistency across embodied tasks execution. 
\begin{wrapfigure}{r}{0.43\textwidth} % {r}表示靠右，0.45\textwidth是图片区域的宽度
    \centering
    \vspace{-0.4cm}
    \includegraphics[width=\linewidth]{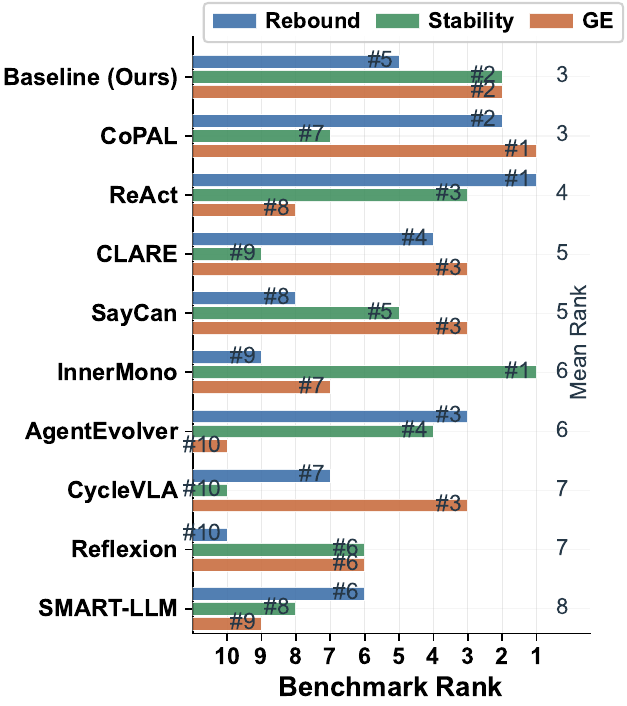} 
    \vspace{-0.7cm}
    \caption{Resilience benchmark landscape.}
    \vspace{-0.7cm}
    \label{fig:resilience_profile_overview}
\end{wrapfigure}

Also, a scalability sensitivity analysis across Qwen3-8B-Instruct, Qwen2.5-7B-Instruct, and Llama3.1-8B-Instruct shows that stronger backbones improve baseline task performance while preserving similar resilience profile patterns, suggesting that resilience is a system-level property rather than only a backbone-size effect (see our \projectresources).

% % 用在RQ1中确定好的指标体系，对各个方法进行韧性指标评估结果，纵向横向对比分析之后，并给出一些具有参考意义的发现
\vspace{-0.3cm}
\subsubsection{Benchmarking EAS Resilience Results}
\vspace{-0.3cm}
\label{sec:experiments_results_rq2}

We apply the validated resilience metrics to benchmark representative EAS methods under identical environments and perturbation settings. Table~\ref{tab:main_benchmark} reports the resilience benchmark scores, and Fig.~\ref{fig:resilience_profile_overview} summarizes the normalized resilience profiles. 
% Methods with similar success rates can exhibit sharply different resilience signatures. 
For example, Reflexion and CoPAL achieve comparable SR (56.3\% vs. 55.8\%), but Reflexion incurs $4.8\times$ higher recovery cost than CoPAL ($57.3$ vs. $11. 9$). This indicates that EAS take different recovery processes, and resilience evaluation reveals these features.

% \subsubsection{Benchmarking Landscape}
% \vspace{-0.4cm}
% \paragraph{Benchmarking Landscape}

% 实验结果统计大表
\begin{table*}[t]
    \centering
    \footnotesize
    \caption{Comparative EAS resilience benchmark results in resilience aspects.} 
    \label{tab:main_benchmark}
    \renewcommand{\arraystretch}{1.1} 
    \setlength{\tabcolsep}{5pt}
    \begin{tabular}{l | c c c | c c | c }
    \toprule
    \multirow{2}{*}{\textbf{Method}} 
    & \multicolumn{3}{c|}{\textbf{Resilience Metrics}} 
    & \multicolumn{2}{c|}{\textbf{Outcome Ref.}}
    & \textbf{Sim.} \\
    & $C_{\mathrm{rec}}$ ($\downarrow$) & $\beta$ ($\downarrow$) & Stress Cap. $\lambda^*$ ($\uparrow$) 
    & \textbf{SR} & \textbf{Comp.} 
    & \textbf{Steps} \\
    \midrule
    \textbf{ReAct}       & 6.5 $\pm$ 0.46  & 0.239 $\pm$ 0.011 & 0.50 $\pm$ 0.042 & 38.5\% $\pm$ 3.1\% & 57.8\% $\pm$ 4.6\% & 3374 $\pm$ 285 \\
    \textbf{Reflexion}   & 57.3 $\pm$ 5.07 & 0.299 $\pm$ 0.021 & 0.66 $\pm$ 0.035 & 56.3\% $\pm$ 4.2\% & 75.0\% $\pm$ 3.8\% & 4390 $\pm$ 312 \\
    \textbf{CycleVLA}    & 35.5 $\pm$ 2.91 & 0.400 $\pm$ 0.026 & 0.72 $\pm$ 0.047 & 22.2\% $\pm$ 2.8\% & 47.5\% $\pm$ 4.1\% & 4688 $\pm$ 420 \\
    \textbf{CLARE}       & 21.6 $\pm$ 1.68 & 0.344 $\pm$ 0.016 & 0.72 $\pm$ 0.052 & 28.4\% $\pm$ 3.5\% & 46.0\% $\pm$ 4.5\% & 4249 $\pm$ 360 \\
    \textbf{SayCan}      & 38.5 $\pm$ 2.49 & 0.283 $\pm$ 0.013 & 0.78 $\pm$ 0.046 & 57.0\% $\pm$ 4.1\% & 72.6\% $\pm$ 3.7\% & 4682 $\pm$ 380 \\
    \textbf{AgentEvolver}& 19.2 $\pm$ 1.24 & 0.264 $\pm$ 0.012 & 0.21 $\pm$ 0.015 & 48.8\% $\pm$ 3.9\% & 66.8\% $\pm$ 4.2\% & 3084 $\pm$ 250 \\
    \textbf{InnerMono}   & 43.5 $\pm$ 2.69 & 0.149 $\pm$ 0.007 & 0.53 $\pm$ 0.026 & 49.5\% $\pm$ 4.0\% & 67.5\% $\pm$ 4.3\% & 3670 $\pm$ 290 \\
    \textbf{CoPAL}       & 11.9 $\pm$ 1.02 & 0.302 $\pm$ 0.017 & 0.85 $\pm$ 0.078 & 55.8\% $\pm$ 4.5\% & 72.6\% $\pm$ 3.9\% & 4846 $\pm$ 410 \\
    \textbf{SMART-LLM}   & 28.2 $\pm$ 2.20 & 0.339 $\pm$ 0.018 & 0.40 $\pm$ 0.036 & 55.7\% $\pm$ 3.8\% & 74.3\% $\pm$ 4.1\% & 5085 $\pm$ 395 \\
    \textbf{Baseline}    & 25.0 $\pm$ 2.03 & 0.200 $\pm$ 0.009 & 0.79 $\pm$ 0.042 & 53.0\% $\pm$ 3.6\% & 69.7\% $\pm$ 3.5\% & 3671 $\pm$ 275 \\
    \bottomrule
    \end{tabular}
    \vspace{-0.2cm}
\end{table*}

% Benchmark 同时说明没有一种方法在我们目前的评测体系中达到最优
This benchmark also shows that no current method dominates all resilience aspects. CoPAL achieves a favorable rebound--extensibility trade-off, with low recovery cost ($C_{\mathrm{rec}}=11.9$) and the highest stress capacity ($\lambda^*=0.85$), but it requires long simulated execution steps. InnerMono has the best stability profile ($\beta=0.149$), yet its recovery cost remains high ($C_{\mathrm{rec}}=43.5$). AgentEvolver has relatively low recovery cost ($C_{\mathrm{rec}}=19.2$), yet its stress capacity is the weakest among all methods ($\lambda^*=0.21$), showing limited graceful extensibility. 
These results demonstrate that resilience should be interpreted as a multi-dimensional execution profile rather than a single scalar ranking.

% \subsubsection{EAS Methods Resilience Signatures} % 横向评述；多个Benchmark的分析，呈现出什么样的特征。
\vspace{-0.4cm}
\noindent \paragraph{EAS Methods Resilience Signatures}
Under the same benchmark protocol, we analyze distinct behaviors (performance, simulation steps, rebound, stability, and graceful extensibility) among the methods in Fig.~\ref{fig:resilience_radar}. The shape of the radar plot serves as a visual fingerprint for the EAS resilience profile, and exhibits different resilience profiles, strength, and weakness. Some methods continue execution through costly recovery, some maintain stable execution, some fluctuate during planning, and some show graceful extensibility under stress. These profiles provide a resilience overview of current EAS methods and serve as a diagnostic handbook for optimization experiments.
We conclude resilience benchmark findings as below:
% Our resilience diagnosis provides a convinent perspective to analyse EAS system resilience in three resilient ways.

\begin{figure}[t]
    % \centering
    \includegraphics[width=\textwidth]{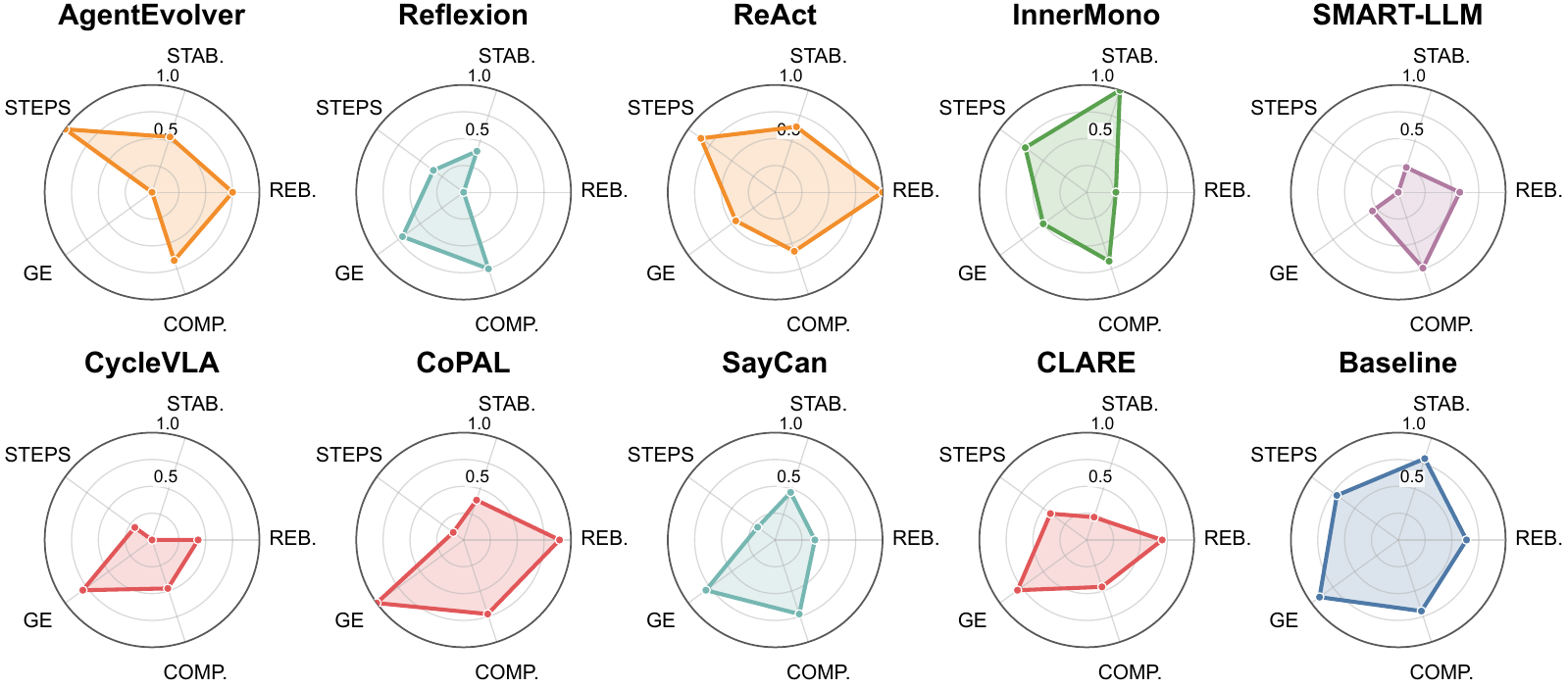} 
    \vspace{-0.2cm}
    \caption{The Resilience Diagnosis. Visualizing the distinct behavioral archetypes of EASs.}
    \vspace{-0.5cm}
    \label{fig:resilience_radar}
\end{figure}

\begin{findingbox}{Resilience Evaluation Benchmark Findings}
\textbf{1) Outcome-equivalent executions can be resilience-inequivalent.}
Success rate and completion describe whether an EAS finishes the task, but they do not reveal how much recovery, replanning instability, or stress margin is consumed during execution. Resilience metrics expose these hidden process differences.

\textbf{2) There is no current EAS method dominates all resilience aspects.}
Some methods recover efficiently, some remain stable under semantic variation, and some tolerate stress, but no method dominates all dimensions. This indicates that EAS deployment should be matched to the target failure regime: difficult tasks require stronger Graceful Extensibility, while sensitive tasks require stronger Stability.

\textbf{3) Resilience metrics provide actionable diagnosis.}
Abnormal metric patterns can be traced back to concrete execution events, such as repeated recovery loops, action oscillation, or stress-boundary collapse. This makes the evaluation layer not only a benchmark tool, but also a diagnostic basis for the metric-guided optimizations in Sec.~\ref{sec:optimization}.
\end{findingbox}

% 这部分内容更多的作为 Practical Handbook，作为 Takeaway
% \begin{findingbox}{Resilience Evaluation Benchmark Findings}
% \textbf{1) Task- and stress-aware selection.}
% A resilient EAS should be selected according to its deployment condition rather than average task success alone. 
% High-stress or resource-missing tasks require stronger Graceful Extensibility, while low-stress but instruction-sensitive tasks require stronger Stability.

% \textbf{2) Cross-aspect trade-offs.}
% Current EAS methods rarely dominate all resilience aspects. 
% A method may recover efficiently but remain unstable, or tolerate higher stress at the cost of larger execution burden. 
% Targeted optimization of one aspect may also affect the others, suggesting that EAS design should be evaluated as a resilience profile rather than a single ranking.

% \textbf{3) Diagnosis-to-optimization.}
% The evaluation layer links abnormal metric patterns to concrete trace events, such as repeated recovery loops, decision oscillations, and stress-margin collapse. 
% These signals support the targeted interventions evaluated in Sec.~\ref{sec:experiments_results_rq3}.
% \end{findingbox}

% ========================================================
\vspace{-0.2cm}
\section{EAS Optimization: Metric-Guided Resilience Diagnostics}
\vspace{-0.2cm}
\label{sec:optimization}
% 对于Rebound层面，我们XXX；对于Stability层面，我们XXX；对于Graceful Extensibility层面，我们XXX
% On rebound aspect, we implement the feedback-guided recovery, which converts environment feedback into recovery hints before retrying. On stability aspect, we add a plan-consistency procedure, which anchors replanning to the current goal, recent actions, and local state to maintain consistency. On graceful extensibility aspect, we design a boundary wait monitor and a cognitive reset mechanism, which mitigates long-tail reasoning and premature collapse in difficult boundary cases.

Based on the resilience metrics and the EAS resilience profiles, we introduce a unified \textit{metric-guided resilience optimization} to the EAS execution. The goal is not to replace the original planner, but to use our resilience metrics as diagnostic signals that indicate where the execution loop should be minimally repaired. We implement three targeted optimizations for Rebound, Stability, and Graceful Extensibility, with implementation details provided in our \projectresources. We compare the \textit{Original} and \textit{Optimized} variants under identical evaluation conditions. Table~\ref{tab:expq3_metric_guided_targeted_improvements} shows that the resilience metrics are not only descriptive, but can also guide practical resilience improvements.
% 值得注意的是，定向优化某单一维度往往会导致其他维度的妥协（例如，优化 GE 导致了 $C_{\mathrm{rec}}$ 的显著增加）。这进一步证实了 Exp-Q2 中的发现：韧性是一种多维度的结构平衡，而我们的指标体系能够精准捕捉这些底层架构的妥协。
What's more, targeted optimization of a single resilient aspect often leads to compromises in other resilience aspects (optimizing GE results in high recovery cost $C_{\mathrm{rec}}$). This confirms our findings: resilience is a multi-dimensional structural balance, and our resilient evaluation layer could capture this trade-off.

% EXP-Q3 性能效果提升对比表
\begin{table}[t!]
% \vspace{-0.1cm}
\centering
\scriptsize
\caption{Comparative results of metric-guided targeted improvements.}
\vspace{-6pt}
% 这里应该加一个注释，说明的只有在进行stress test的情况下，GE的斜率等信息才具有意义
\label{tab:expq3_metric_guided_targeted_improvements}
\renewcommand{\arraystretch}{1.12}
\setlength{\tabcolsep}{2.5pt}
\setlength{\arrayrulewidth}{0.35pt}
\begin{tabular}{l|l|c c c c c|c}
\toprule
\textbf{Target}
& \textbf{Methods}
& \textbf{Rebound $C_{\mathrm{rec}}$} $\downarrow$ 
& \textbf{Rec. Win.} $\downarrow$ 
& \textbf{$\beta_{\mathrm{step}}$} $\downarrow$ 
& \textbf{$\beta$} $\downarrow$ 
& \textbf{GE. Comp.} $\uparrow$ 
& \textbf{Formal GE} \\
\midrule
 & Baseline & 59.11 & 3.13 & 0.2881 & 0.6834 & 0.887 & -- \\
\rowcolor{cyan!9}
\cellcolor{white}\multirow{-2}{*}{\textbf{Rebound}} & \textbf{Optimized} & \underline{\textbf{33.73}} {\scriptsize\textcolor{teal!70!black}{$\downarrow$42.94\%}} & \underline{\textbf{2.38}} {\scriptsize\textcolor{teal!70!black}{$\downarrow$23.96\%}} & 0.2079 {\scriptsize\textcolor{teal!70!black}{$\downarrow$27.84\%}} & 0.6870 {\scriptsize\textcolor{red!70!black}{$\uparrow$0.53\%}} & 0.819 {\scriptsize\textcolor{red!70!black}{$\downarrow$7.67\%}} & -- \\
\midrule
 & Baseline & 51.01 & 2.63 & 0.3185 & 0.6788 & 0.903 & -- \\
\rowcolor{cyan!9}
\cellcolor{white}\multirow{-2}{*}{\textbf{Stability}} & \textbf{Optimized} & 43.01 {\scriptsize\textcolor{teal!70!black}{$\downarrow$15.68\%}} & 3.00 {\scriptsize\textcolor{red!70!black}{$\uparrow$14.07\%}} & \underline{\textbf{0.2552}} {\scriptsize\textcolor{teal!70!black}{$\downarrow$19.87\%}} & \underline{\textbf{0.6772}} {\scriptsize\textcolor{teal!70!black}{$\downarrow$0.24\%}} & 0.750 {\scriptsize\textcolor{red!70!black}{$\downarrow$16.94\%}} & -- \\
\midrule
 & Baseline & 36.70 & 1.33 & 1.4594 & 0.6917 & 0.833 & Incomplete \\
\rowcolor{cyan!9}
\cellcolor{white}\multirow{-2}{*}{\textbf{GE}} & \textbf{Optimized} & 99.59 {\scriptsize\textcolor{red!70!black}{$\uparrow$171.36\%}} & 2.25 {\scriptsize\textcolor{red!70!black}{$\uparrow$69.17\%}} & 0.7215 {\scriptsize\textcolor{teal!70!black}{$\downarrow$50.56\%}} & 0.6371 {\scriptsize\textcolor{teal!70!black}{$\downarrow$7.89\%}} & \underline{\textbf{0.917}} {\scriptsize\textcolor{teal!70!black}{$\uparrow$10.08\%}} & Complete$^{\dagger}$ \\
\bottomrule
\multicolumn{8}{l}{\footnotesize $^{\dagger}$ Formal GE is reported only for GE stress-tests, since it requires a stress-response curve over $\lambda$ grid.} 
\end{tabular}
\vspace{-0.4cm}
\end{table}

\vspace{-0.45cm}
\noindent \paragraph{Opt1 Recovery: Feedback-Guided Recovery Implementation}
High recovery cost $C_{\mathrm{rec}}$ and long recovery windows indicate that the agent often spends extra reasoning or physical actions after local failures. To address this failure mode, we add a feedback-guided recovery layer on top of the centralized LLM planner. 
The layer monitors World-Graph evidence, recent execution feedback, and task-progress stagnation, and diagnoses recoverable faults such as missing graph nodes, stale object locations, repeated tool failures, or invalid object references. 
Instead of overriding the planner with a separate recovery policy, we get \textit{Rebound Guidance} $\Pi_{t+1}$ into the next planning context, including perception refresh, reflection, state summarization, or belief rollback. Then we formalize it as Habitat skill tools for better use. This keeps the original action space unchanged while converting low-level execution feedback $\mathcal{F}_t$ into planner recovery hints. As shown in Table~\ref{tab:expq3_metric_guided_targeted_improvements}, this optimization reduces $C_{\mathrm{rec}}$ from $59.11$ to $33.73$ and the recovery window from $3.13$ to $2.38$, showing that the metric-guided feedback loop directly reduces redundant recovery effort.

\vspace{-0.45cm}
\noindent \paragraph{Opt2 Stability: Consistency State Record}
High $\beta$ values reveal unstable replanning behavior, where the agent adapts to new feedback but may oscillate between similar actions, repeat completed subtasks, or drift away from the current execution $c_t$. 
We therefore introduce a consistency record for replanning. At each step, it constructs a read-only phase guidance from the task goal, current phase, World-Graph state, current room and holding state, completed and pending subtasks, and recent action-response history. This design improves consistency without making the policy brittle. The optimized variant reduces $\beta_{\mathrm{step}}$ from $0.3185$ to $0.2552$ and improves $\beta$ from $0.6788$ to $0.6772$, indicating that the stability metric captures and guides reductions in local replanning variance.

\vspace{-0.45cm}
\noindent \paragraph{Opt3 GE: Boundary Long-tail Control for Difficulties}
Graceful Extensibility focuses on whether performance degrades smoothly as stress increases, rather than collapsing or entering unbounded long-tail behavior. In high-stress cases, we observed that the planner may repeatedly wait or receive near-duplicate execution feedback without making progress. Such behavior inflates execution cost and makes the stress-response curve insensitive to true adaptive capacity. 
To control this boundary effect, we add a long-tail monitor that detects dual-wait patterns, semantically repeated feedback, and stagnant task progress. When a boundary stall $B_t$ is detected, the system injects a bounded cognitive reset that discourages further non-progressive waiting and requests either a concrete productive action or termination. This converts uncontrolled long-tail execution into measurable boundary execution.

% ========================================================
\vspace{-0.4cm}
\section{Conclusion and Future Work}
\label{sec:conclusion_future}
\vspace{-0.3cm}
We presented the first Resilience Evaluation Framework for EASs. By designing novel metrics: \textit{Rebound}, \textit{Stability}, and \textit{Graceful Extensibility}, our framework can expose resilience hidden pathologies, such as high-cost recovery and catastrophic forgetting, which remain invisible to outcome-centric metrics. Empirical analysis across representative baselines identified distinct Resilience Archetypes, confirming that architectural choices involve quantifiable trade-offs between execution, stability, and adaptability. The proposed optimization framework can help EASs developers to diagnose root causes and enforce non-degradation contracts to build a resilient and trustworthy EAS.
% We implemented a process-aware \textbf{Resilience Evaluation Layer} for EAS evaluation complement, moving evaluation outcomes toward execution-level resilience analysis. 
% Grounded in resilience engineering, we measured three resilience metrics aspects: \textit{Rebound} through Recovery Cost $C_{\mathrm{rec}}$, \textit{Stability} through $\beta$-Stability, and \textit{Graceful Extensibility} through the stress-response mapping $M_f(\lambda)$, stress capacity $\lambda_f^*$. 
% These metrics quantify how an EAS recovers from disruption, remains consistent under small perturbations, and preserves acceptable execution as stress increases.

% \vspace{-0.1cm}
\noindent\textbf{Future Work}
% Sim2Real, Theoretical Trade-off Analysis, Practical Optimization based on EAS Deployments
Future work can extend resilience evaluation along three directions: \textit{Sim2Real} validation of whether simulation-derived resilience profiles transfer to physical robots, \textit{theoretical analysis of trade-offs} among resilience aspects, and \textit{deployment-aware optimization} that uses benchmarked resilience signatures to improve EASs for target application conditions. These directions can move resilience evaluation toward more principled, deployable, and balanced EAS design.
% Together, these directions would move resilience evaluation toward principled, deployable, and balanced EAS design.


% \begin{acks}
% \end{acks}
  
% \clearpage
% \bibliographystyle{ACM-Reference-Format}
\bibliographystyle{plain}
\bibliography{references}

\end{document}